\documentclass[12pt]{article} 
\usepackage{amsmath,amssymb,amsthm}
\usepackage{cite}
\usepackage[usenames]{xcolor}
\usepackage{comment}
\usepackage[shortlabels]{enumitem}
  \setlist[enumerate,1]{leftmargin=20pt,nosep}
\usepackage[paper=a4paper]{geometry}
\usepackage{graphicx}
\usepackage{hyperref}
\usepackage{multicol}
\usepackage[vskip=0pt,leftmargin=24pt]{quoting}
\usepackage{wasysym}
\usepackage{xurl}
\newcommand\Ar{\smallskip\noindent\texttt{A:\ }}

\newcommand\GD{\ensuremath{\mathrm{GD}^\theta_{T,C}}}

\renewcommand\phi{\varphi}
\newcommand{\Qn}{\smallskip\noindent\texttt{Q:\ }}

\title{On a measure of intelligence}
\author{Yuri Gurevich}
\date{}
\begin{document}
\maketitle
\thispagestyle{empty}

\begin{quoting}\raggedleft\it
The measure of intelligence is the ability to change.\\[1ex]
--- Albert Einstein\footnote%
{The saying is commonly attributed to Einstein.
A similar saying ``Intelligence is the ability to adapt to change''  is commonly attributed to Steven Hawking.
In neither case is there a clear reference. }
\end{quoting}
\raggedright

\begin{abstract}
The Fall 2024 Logic in Computer Science column of the Bulletin of EATCS is a little discussion on intelligence, measuring intelligence, and related issues, provoked by a fascinating must-read article ``On the measure of intelligence'' by Fran\c{c}ois Chollet.
The discussion includes a modicum of critique of the article.

\end{abstract}

\subsection{Cybernetics vs.\ AI, 
and podcasts vs.\ reading}
\noindent\texttt{Quisani\footnote{My former student}} {\small(walking in)}:\  
What are you reading?

\smallskip\noindent\texttt{Author:} 
An article ``On the measure of intelligence'' by Fran\c{c}ois Chollet \cite{Chollet2019}.

\Qn Is it about psychology?

\Ar It is mostly about AI.
Chollet is a prominent figure in AI.

\Qn We spoke about AI last spring. 
But you didn't seem to be interested in AI before that.

\Ar This is largely correct, though I read Norbert Wiener's  ``Cybernetics'' \cite{Wiener}, when it was translated to Russian in 1968,  and was taken with it.
For a while I tried to follow cybernetics developments, at least in the USSR. 

\Qn What's cybernetics?

\Ar Wiener gives a concise definition in the subtitle of that book of his: ``Control and communication in the animal and the machine.'' 

\Qn How is this different from AI? 

\Ar This is a good question.
Here is an explanation by Michael Jordan, UC Berkeley Professor, not the basketball player: 
\begin{quoting}
It was John McCarthy (while a professor at Dartmouth, and soon to take a position at MIT) who coined the term AI, apparently to distinguish his budding research agenda from that of Norbert Wiener (then an older professor at MIT). Wiener had coined ``cybernetics'' to refer to his own vision of intelligent systems --- a vision that was closely tied to operations research, statistics, pattern recognition, information theory, and control theory. McCarthy, on the other hand, emphasized the ties to logic. In an interesting reversal, it is Wiener’s intellectual agenda that has come to dominate in the current era, under the banner of McCarthy’s terminology \cite{Jordan2019}.
\end{quoting}
In his conversation with Lex Fridman, Jordan tells this story in more colorful terms \cite[18:06]{Jordan2020}.

In the USSR of my time (I left ``the land of victorious socialism'' in 1973) and for a long time after, the term cybernetics was used for the field.

\Qn Do you spend much time reading stuff outside your immediate research interests?

\Ar I do, though these days I spend more time listening to podcasts.

\Qn Why?

\Ar You can walk and listen to a podcast; you can't walk and read.

\Qn I can't but some do, even as they cross the road. But I digress. 

\Ar A podcast often provokes reading.
Something attracts your attention, and you want to know more about it. 
That is exactly what happened in this case. 
First I heard Fran\c{c}ois Chollet on a podcast \cite{Chollet2020}.

\subsection{The g-factor}

\Qn Is Chollet's article about IQ scores? 
I have never heard of any other measure of intelligence. 

\newpage
\Ar Chollet mentions IQ scores and the g-factor, but his 64-page article goes far beyond that.

\Qn What's the g-factor?

\Ar British psychologist Charles Spearman noticed positive correlations when a person performs different tests.
He suggested that general intelligence (g-factor) is a single
underlying ability influencing (but not determining) performance across different cognitive tasks \cite{Spearman}.

\Qn I know people who perform well on one kind of problems, say complex word problems, but not so well on another kind of problems, say mathematical problems.

\Ar Richard Haier, the editor-in-chief of \emph{Intelligence}, says that, typically,  the correlation is still positive \cite[04:32] {Haier}.

\Qn How well has Spearman's idea held up?

\Ar There is an ongoing debate about its validity and limitations, especially in modern psychology, but it seems to be accepted in prinicple.
``I think I can say without fear of being empirically contradicted that it is the most replicated
finding in all of psychology'' \cite[06:05]{Haier}.

\Qn Wow! And IQ scores are supposed to measure the g-factor?

\Ar Modern IQ tests are designed to do just that by assessing various cognitive domains.

\Qn It is hard to believe that intelligence, that is general intelligence, can be summarized by one number. 

\Ar I agree with you. Intelligence is too complex and multifaceted to be reduced to one number.
But the proponents of IQ scores point out its empirical support, practicality, and predictive utility.
This issue is too involved and important to be discussed on one foot, as they say in Hebrew.
We would need to give it the time and attention it deserves.

\Qn I understand.

\subsection{Intelligence as skill-acquisition efficiency}

\Qn I guess Chollet also wants to measure the g-factor. 
What is special about his approach?

\Ar To make progress towards the promise of AI, one needs a workable definition of intelligence and a quantitative measure of intelligence – in particular human-like general intelligence.
Chollet makes a good step in this direction.
\begin{quoting}
We \dots\ articulate a new formal deﬁnition of intelligence based
on Algorithmic Information Theory, describing intelligence as \emph{skill-acquisition efﬁciency} and highlighting the concepts of \emph{scope, generalization difﬁculty, priors, and experience}, as critical pieces to be accounted for in characterizing intelligent systems \cite[Abstract]{Chollet2019}.
\end{quoting}
A task-specific performance, say at chess or Go, measures a particular skill; the performance can be inflated through training data and prior knowledge.
Whether you are a human or an AI system,  general intelligence is not about what you know or can do. 
According to Chollet, it is about how efficiently you can acquire a new skill or adapt to a new environment that you did not anticipate.
``There's a big distinction to be drawn between intelligence, which is a process, and the output of that process
which is skill'' \cite[28:52]{Chollet2019Chat}.

\Qn I always thought about intelligence as a capacity, rather than a process. 
Maybe Chollet means ``skill-acquisition'' which is a process. 
In any case, I like very much the definition of intelligence as skill-acquisition efficiency. 
What about you?

\Ar I doubt that the sprawling intuitive notion of intelligence reduces to skill-acquisition efﬁciency but it is hard to argue with the thesis that skill-acquisition efﬁciency is a manifestation of intelligence.

\Qn What does efficiency mean in Chollet's definition?
Humans can often grasp a new concept with just a few examples. 
Is that the kind of efficiency he is talking about?

\Ar Yes, an efficient system can generalize from limited examples to solve a broader range of related problems.
This is an important aspect of skill-acquisition efficiency.
But there are other aspects. In particular, an efficient system can
quickly adapt to moderate changes in its environment without extensive retraining, can apply knowledge from previously learned tasks to a new one, and can use less compute.

\subsection{Algorithmic Information Theory (AIT)}

\Qn It is surprising that skill-acquisition efficiency has a formal definition.
Does Chollet prove interesting theorems about it?

\newpage
\Ar Chollet expresses various notions involved in his definition of intelligence in the AIT formalism. 
He proves no theorems.
To me, the formalization exercise is not convincing. 
It is not even sound in a sense.

\Qn What do you mean?

\Ar Let me introduce Chollet's notation and, at the same time, recall algorithmic complexity, also known as Kolmogorov complexity \mbox{}\cite{LV,Shen}.
Below, strings are binary strings.
\begin{quoting}
The Algorithmic Complexity (noted $H(s)$) of a string $s$ is \dots\ the length of the shortest program that outputs the string when running on a ﬁxed universal Turing machine. Since
any universal Turing machine can emulate any other universal Turing machine, $H(s)$ is machine-independent to a constant.
\dots [D]eﬁne the information content that a string $s_2$
possesses about a string $s_1$ (called “Relative Algorithmic Complexity” and noted $H(s_1|s_2)$)
as the length of the shortest program that, taking $s_2$ as input, produces $s_1$
\cite[p.~34]{Chollet2019}. 
\end{quoting}
Machine-independence follows from the fact that a given universal programming language $L$ can emulate any universal programming language $L'$ (as well as any non-universal programming language) up to an additive constant.
As a result, $H_L(s) \le H_{L'}(s) + c$ and $H_L(s_1|s_2) \le H_{L'}(s_1|s_2) + c$, where $c$ is essentially the length of an $L'$-to-$L$ translator.
If machine independence is a must, then Chollet has a problem. 
For example, he formalizes the \emph{generalization difficulty} \GD\ of a task $T$ given an experience curriculum $C$ as 
\begin{equation}
\frac{H(\mathrm{Sol}^\theta_T | \mathrm{TrainSol}^{opt}_{T,C})} {H(\mathrm{Sol}^\theta_T) }
\tag{GD}
\end{equation}
where $\mathrm{Sol}^\theta_T$ is the shortest solution of $T$ of threshold $\theta$ during evaluation, and $\mathrm{TrainSol}^{opt}_{T,C}$ is the shortest optimal training-time solution of $T$ given $C$, so that $0<  \GD \le1$.
(Chollet describes $T, C, \theta, \mathrm{Sol}^\theta_T $, and $\mathrm{TrainSol}^{opt}_{T,C}$ more precisely, but this is  not important for our purposes.)
The allegedly innocent additive constant makes \GD\ highly dependent on the fixed universal language, not just up to an additive constant.

\Qn Explain.

\Ar Notice that the additive constant $c$ may be arbitrarily large. 
Indeed, flip a fair coin $N$ times, thus producing a random string $s$ of length $N$.
Then, with overwhelming probability, $H(s)\approx N$ for the fixed language $L$ \cite[p.~8]{Shen}.
If $L'$ is the extension of $L$ with a single short command that prints $s$ then, for $L'$, $H(s)$ is a small constant, and so $c\approx N$.

Let $A$ and $B$ be the numerator and denominator in (GD).
Consider a situation where, for our fixed language $L$, \GD\ is small, say $A = 1$ and $B = 100$, so that GD$^\theta_{T,C} = 1/100$.
Let $L'$ be another language whose Kolmogorov complexities exceed those of $L$ by $1 000 000$. 
Then, for $L'$, we have $A = 1 000 001$ and $B = 1 000 100$, so that GD$^\theta_{T,C}\approx 1$, so that  \GD\ moves from one extreme ($\approx0$) to the other ($\approx1$).

\Qn What does curriculum mean in Chollet's approach?

\Ar A \emph{curriculum} is a (carefully crafted) sequence of tasks. 
It is an important concept in Chollet's approach.
Curricula aim to cover a wide range of task-solving skills and are designed to minimize the reliance on pre-existing knowledge.

\Qn Back to your critique, while machine independence is desirable, one still can work with a fixed universal programming language, carefully chosen to minimize built-in information.

\Ar This is true, at least in principle. 
But let me notice that the functions $H(s)$ and $H(s_1|s_2)$ are uncomputable. 
Moreover, they are not even approximable by computable functions.
Indeed suppose toward a contradiction that $|H(s) - f(s)| \le N$ for some computable $f(s)$.
Then $f(s) - N$ is a computable lower bound for $H(s)$.
By Theorem~6 in \cite{Shen}, $f(s) - N$ is bounded.
Then $f(s)$ is bounded and therefore $H(s)$ is bounded, which is impossible.

\Qn Still, there may be something useful in algorithmic complexity. 
An expression like (GD) is more succinct than a textual description of it. 

\Ar Yes, symbolic expressions may be succinct, and this is an advantage. 

\Qn How important is the AIT foundation for Chollet's approach?

\Ar  I don't think it is important. 
Chollet's approach isn't really founded in AIT. 
In the literature, AIT is typically used as motivation and inspiration rather than literally, and it can play such a role in Chollet's approach as well.

\subsection{Measuring intelligence}

\Qn What else is there in Chollet's article?

\Ar He covers a lot of ground. 
There is a general theme of defining intelligence, that we discussed above, and measuring intelligence, that I intend to bring up. 
But many sections are educational essays in their own right, at least for non-experts.
For example, there is a section \S I.3.2 on generalization theory.
It addresses deﬁning, measuring, and maximizing generalization.

\Qn Advanced mathematics is all about generalizations.
But I have never heard of measuring generalization.
Maximizing generalization would probably result in some trivial scenario, with no interesting theorems.

\Ar Chollet speaks about different notion of generalization, ``originally developed to characterize how well a statistical model performs on inputs that were not part of its training data'' \cite[p.~9]{Chollet2019}.


%
Measuring intelligence is super important to Chollet: ``we need to be able to deﬁne and evaluate intelligence in a way that enables comparisons between two [AI] systems, as well as comparisons with humans'' \cite[Abstract]{Chollet2019}.
To this end, he reifies his definition of intelligence as ``skill-acquisition efﬁciency over a scope of tasks, with respect to priors, experience, and generalization difﬁculty'' \cite[p.~27]{Chollet2019}.

\Qn I have encountered the noun prior before, but only in Bayesian statistics, where it means prior distribution. 

\Ar In AI the noun prior, often in the form priors, is used to mean  previously acquired knowledge that the agent, an AI system or a human, brings to the table.
Chollet proposes a new benchmark for general intelligence that he calls Abstraction and Reasoning Corpus (ARC). 

\Qn How does ARC differ from previous benchmarks?

\Ar Traditionally, AI benchmarks focused on tasks solvable through pattern recognition and statistical learning.
ARC emphasizes generalization, abstraction, and few-shot learning.

\Qn I haven't heard of few-shot learning but the name seems to suggest learning new tasks from a small number of examples.

\Ar Exactly. 

\Qn ARC seems ambitious.

\Ar It is very ambitious, a paradigm shift in how we assess intelligence. 
ARC aims to provide a standard measure for generalization and abstraction abilities of different intelligent systems, whether AI systems or humans. 

\Qn Is ARC already available? Are there some tests that I can try?

\Ar Yes, see \cite{Arc}. 
There are quite a number of tests.

\Qn Great. I intend to try some of those tests, at least on myself.

\Ar You will need a GitHub account and a bit of programming prowess to make those tests available in convenient visual form. 

\Qn ARC looks super attractive but also super challenging, especially if one wants to compare AI systems with humans.

\Ar I agree. The project is grandiose, and challenges are abundant.
In particular, ARC involves new algorithms to assess abstract reasoning and few-shot learning. 

\Qn Does Chollet propose ARC as an adequate measure of intelligence?

\Ar Not at this point. ARC is a project in progress. 
The current implementation presupposes only human core knowledge and even that only in part.
One has to start somewhere.

\Qn What's core knowledge?

\Ar The knowledge that we are born with or hardwired to acquire quickly after birth. 

\Qn To compare AI systems with humans, Chollet needs to make human core knowledge explicit. 
Is this even possible?

\Ar There has been impressive progress on understanding the human core knowledge \cite{Spelke}.
The core knowledge comprises a few distinct systems, i.e.\ distinct priors. 
One prior allows us to see the world split into distinct objects. 
Another prior is that some of these objects are goal-pursuing agents. 
Two additional priors are related to primitive geometry and numbering respectively. 
``Human cognition is founded, in part, on four systems for representing objects, actions, number, and space. It may be based, as well, on a ﬁfth system for representing social partners'' \cite[Abstract]{Spelke}. 
%
%
%

\subsection{Reaction to Chollet's paper}

Q: Has Chollet’s definition of intelligence been accepted?

A: It generated significant debate in the AI research community.
It influenced discussions on AI evaluation metrics and benchmarks. 
It sparked renewed interest in developing more comprehensive intelligence tests for AI systems.
But it's not universally accepted.
Some question whether Chollet's definition of intelligence is comprehensive. 
Does it capture creativity? 
Does it capture emotional intelligence?

\newpage
\Qn I understand creativity doubts.
As far as emotional intelligence is concerned, my impression is that it is irrelevant here. 
That is unless advanced intelligence is impossible without emotional intelligence.

\Ar Recent advances in neuroscience shed light on this question. 
Jeff Hawkins, neuroscientist and businessman, addresses the issue head-on in his beautiful book  The Thousand Brains Theory of Intelligence \cite{Hawkins}.

The parts of the brain responsible for intelligence and emotion are distinct, albeit connected.
\begin{quoting}
The newest part of our brain is the neocortex \dots\
The neocortex is the organ of intelligence. \dots\
Fears and emotions are created by neurons in the old brain \cite[\S1]{Hawkins}.
\end{quoting}
Emotionless artificial intelligence is not only possible; it is more straightforward.
\begin{quoting}
Intelligent machines need to have a model of the world and the
flexibility of behavior that comes from that model, but they don’t need to have human-like instincts for survival and procreation \cite[\S10]{Hawkins}.\dots\\
We, the designers of intelligent machines, have to go out of our way to design in motivations \cite[\S11]{Hawkins}.
\end{quoting}

\Qn Should we design in emotions?

\Ar Personally, I don't think so. 
Can you switch off a computer that implores you not to do that?
But the issue is deep.

\Qn I am becoming a serial digresser \smiley{}
Let's return to Chollet.
Since he has the ambition to compare AI with humans, what do neuroscientists say? What, if anything, do computer scientists say?

\Ar I don't have a representative sample of opinions. 
My impression is that, by and large, neuroscientists thought and continue to think that intelligence is poorly defined. Recall Michael Jordan whom I quoted earlier. He says: “We don’t know what intelligence is. We don’t know much about abstraction and
reasoning at the level of humans; we don’t have a clue” [9, 16:28].

Computer scientist Leslie Valiant, who received the Turing Award
in 2010, has a similar opinion, with a twist.
In his conversation with Sean Carroll \cite{ValiantChat} he says:
\newpage
\begin{quoting}
I think the main downside of intelligence is that no one can define it. That is, of course, people have complained that we give importance to intelligence. 
We test people for intelligence, and this has consequences.
And we don’t even know what we’re testing for, where the questions come from. So I think it’s very unfortunate that the notion of intelligence has become so important, because it’s not explicitly defined. So I’ve explicitly defined educability; intelligence has no such definition [14, 45:51].
\end{quoting}
Q: I see what you mean by the twist. Tell me about educability.

A: I will, though not now. Valiant’s approach [13] deserves a deep dive, a separate
conversation.

Q: OK, I understand. I wonder whether Michael Jordan and Leslie Valiant heard of Chollet's definition of intelligence and what they think about it.

A: I asked Valiant about that.
Here's his reply:
\begin{quoting}
I define a model (educability) and describe what real world phenomenon it is intended to correspond to (extra cognitive capabilities of humans).
Chollet does indeed also have a model. But my question is: what real world phenomenon does it correspond to? Intelligence as generally used is a broad sprawling concept but some aspects, such as what IQ tests measure, seem inseparable from it. I think it is difficult to argue that any model of intelligence corresponds well to the intuitive concept as widely
used. I think the challenge for any model is to articulate what important phenomenon it captures [15].
\end{quoting}

\subsection{Logic}

\Qn We can't finish without touching on logic.
I am looking at the official site of Abstraction and Reasoning Corpus for Artificial General Intelligence \cite{Arc}.
The terms abstraction and reasoning should make the logician in you happy.

\Ar You bet. 
But such abstractions and reasoning are a challenge to the logic community.

\Qn Yes, we spoke about this last spring \cite{G258}.

\Ar  Let me add a few words. 
Many mathematicians and mathematical logicians look down on philosophers. 
There is a joke about the chairman of math department trying to squeeze out a new position from the dean:
\newpage
\begin{enumerate}
\item[---\ ] Contrary to physicists and chemists, mathematicians don't need labs, just a pen, paper, and a waste basket.
\item[---\ ] You think you are cheap. The chairman of the philosophy department tells me that philosophers need only a pen and paper, no waste basket.
\end{enumerate}

\Qn That's cruel.
Also, philosophical departments don't have a monopoly on philosophy. 

\Ar I agree on both counts. In any case, mathematical logicians should not forget that it was essentially philosophical work of formalization that prepared the ground for mathematical logic. 
Think of Boole, Cantor, Frege, Russell  \&\ Whitehead, Zermelo. 
The logics of today are not sufficient to satisfy the needs of rapidly developing AI. 

\Qn Yes, in our conversation last spring, you mentioned fast thinking in the sense of Kahneman \cite{Kahneman}. 

\Ar Right. We need new foundational investigations, new formalizations which would enable us to develop logics appropriate for new applications.

\Qn Who will do the work?
Mathematical logicians or philosophers?
Or maybe computer scientists?

\Ar The work requires (at least) intimate knowledge of AI, mathematical maturity, and a philosophical/foundational attitude. 
An appropriate person can start his career in AI, computer science, logic, mathematics,  philosophy, or some different field altogether.

\Qn It could be a team.

\Ar Yes, it could be a team. Russell and Whitehead worked as a team, with a clear division of labor.
Daniel Kahneman and Amos Tversky did their foundational work, albeit on psychology rather than logic, as a closely knit team \cite{Kahneman}

\subsubsection*{Acknowledgments}
Many thanks to Andreas Blass and Alexander Shen for generously and repeatedly commenting on drafts of the paper.

\end{document}